\documentclass[pdflatex,sn-nature]{sn-jnl}

\usepackage{graphicx}%
\usepackage{multirow}%
\usepackage{amsmath,amssymb,amsfonts}%
\usepackage{amsthm}%
\usepackage{mathrsfs}%
\usepackage[title]{appendix}%
\usepackage{xcolor}%
\usepackage{textcomp}%
\usepackage{manyfoot}%
\usepackage{booktabs}%
\usepackage{algorithm}%
\usepackage{algorithmicx}%
\usepackage{algpseudocode}%
\usepackage{listings}%
\usepackage{amsmath}

\theoremstyle{thmstyleone}%
\theoremstyle{thmstyletwo}%

\theoremstyle{thmstylethree}%

\usepackage{setspace}
\setcitestyle{super}
\begin{document}

\title[Article Title]{Communications-free Collective Intelligence Via Allocentric Belief}


\author*[1]{\fnm{Cameron} \sur{Mehlman}}\email{cpm222@cornell.edu}

\author[1]{\fnm{Arjun} \sur{Mulchandani}}\email{abm254@cornell.edu}

\author[1]{\fnm{Hovik} \sur{Grigoryan}}\email{hg483@cornell.edu}

\author[1]{\fnm{Anurag} \sur{Koyyada}}\email{ak2542@cornell.edu}

\author[1]{\fnm{Gregory} \sur{Falco}}\email{gfalco@cornell.edu}

\affil*[1]{\orgdiv{Sibley School of Mechanical and Aerospace Engineering}, \orgname{Cornell University}, \orgaddress{\city{Ithaca}, \postcode{14850}, \state{NY}, \country{United States}}}

\abstract{A team of skilled card players share no information amongst each other, and yet still are able to coordinate their play by watching the same players and reading the same tells. Autonomous systems have failed to exploit this logic, relying instead on explicit communication \cite{1435481,balch1994communication,31570963157348,7759822} or complex state estimation schemes \cite{indelman2018cooperative,7759822,ZHANG2022341,fang2023factbasedagentmodelingmultiagent,9534358} that can fail when it matters most. Without it, decentralized agents diverge in their world beliefs, and coordination collapses. Here we show that anchoring all agents to a single shared external actor, a common witness whose reactions encode team state, produces spontaneous, communication-free coordination. An allocentric belief network trained on witness behavior allows each agent to independently reconstruct full team state, keeping group beliefs consistent with no inter-agent exchange. Across five competitive environments allocentric belief alignment outperforms egocentric approaches \cite{10333311,ito2026robottask,krogius2021decentralized} in collective performance, and even scales to sixty-agent teams where standard methods fail. These results establish a new principle for collective intelligence: in the absence of communication, a shared gaze replaces a shared channel.}



\maketitle

\section{Introduction}\label{sec:intro}

Teams of robots that cannot talk to one another such as drones beyond radio range, or swarms in remote, and undeveloped regions seem limited to what each member can see. Yet hunters coordinate around prey without speaking, because the prey's reaction tells each hunter what the others know. This is not dissimilar to poker -- a popular application in early game theory works \cite{nash1950simple,von2007theory} -- where a skilled player can learn to read another's tells.
And, although these works did not assume inferential observation they laid the foundation for multi-agent game theory which has since taken root in numerous fields across science and engineering including decentralized 
coordination for collective behavior \cite{durfee88, zhang13,10.3389/frobt.2018.00059,kurach2020google, 9201368, shirado2017locally, van1990tacit, mehta1994nature}. We study a largely unsolved problem arising from this line of research: how can decentralized agents coordinate without communication and with limited sensing? We investigate whether decentralized agents can estimate task-relevant team state by observing a single external actor whose behavior responds to the team. We refer to this actor as a witness. Common witness equilibrium examines how team response policies shape an opponent’s incentives to reveal or conceal information\cite{falco2026witness}. Here, we develop learned state estimators and evaluate decentralized control with fixed witness policies across robotic and economic environments.

Decentralized collaborative control in communications-denied environments poses many challenges, but perhaps the greatest obstacle to coordination is severely limited sensing -- often modeled as a single field-of-view (FOV) constraint \cite{11127700,10333311,8901066,9698245} -- in which each agent can observe only one environment entity at a time, and complicates group coordination. Furthermore, a common witness outside the team can compound environment complexity by introducing dynamics that can make teammates’ movements incredibly difficult to predict. In order to achieve stable decentralized team behavior in these highly dynamic and complex systems we develop a decentralized Reinforcement Learning (RL) training framework that leverages a centralized expert to learn a decentralized policy. This provides us with a baseline behavior to expect from the multi-agent system, however, the decentralized policies still react from global state. The problem at hand then becomes one of state estimation: how can agents that each observe only a small part of the environment at any timestep maintain consistent and accurate estimates of the global state? Our approach trains each agent to estimate task-relevant team state from a history of observations of the same external actor.

Traditionally, decentralized cooperative teams in partially observable environments attempt to fill the gaps in unobserved environment through state estimation, however, often suffer from the same failure mode: divergence in believed state among agents degrades coordination. Explicit state estimation methods for group coordination such as communications based data-fusion \cite{1435481,balch1994communication,3157096.3157348}, and model-based estimation of unobserved environment variables \cite{indelman2018cooperative,7759822,ZHANG2022341,fang2023factbasedagentmodelingmultiagent,9534358} are plagued by this due to the asynchronicity of individual team members observations. Other methods that support implicitly estimating hidden environment variables from limited observation such as inference learning
\cite{Sukthankar-2014-17164,Li2018DynamicAA,bardnolarn,pmlr-v48-he16,pmlr-v80-raileanu18a, 10536732}, and stigmergy 
\cite{salman2024automatic,doi:10.1126/science.1245842,bonabeau1999swarm, 6788132} are often employed to predict more general environment phases or interpret simple passive team messages, and cannot provide rich, full state estimates. As a result, no existing approach reliably provides decentralized cooperative agents with consistent, complete estimates of the global state.

\begin{figure}[h]
\centering
\includegraphics[width=0.7\textwidth]{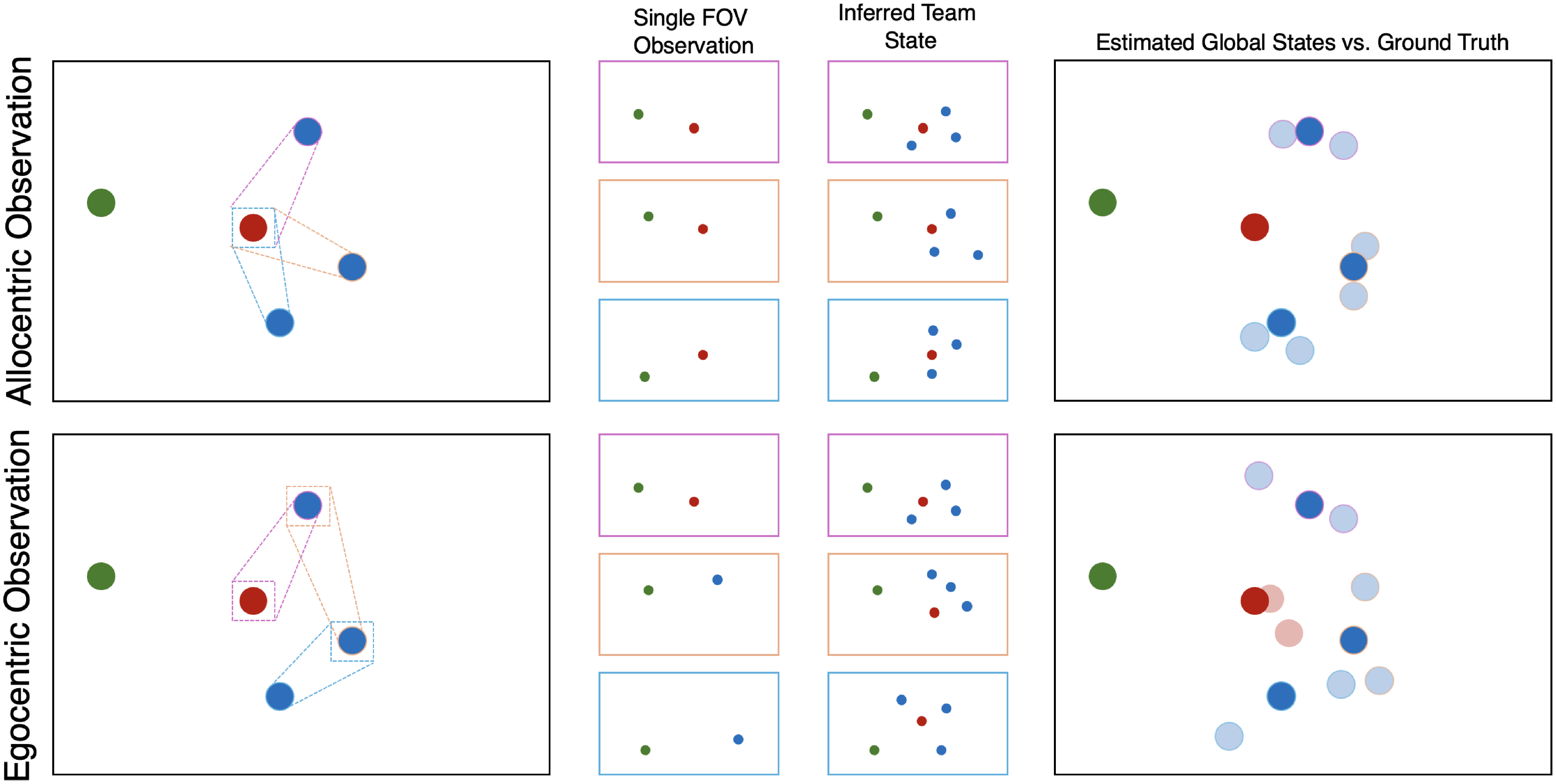}
\caption{Allocentric versus egocentric state estimation, and our framework. With allocentric observation (top), all agents observe the same non-cooperative witness, keeping their state estimates consistent. Conversely, with egocentric observation (bottom), agents observe different features asynchronously, and the variance in their global state estimates grows over time.}\label{fig:master_intro}
\end{figure}

We address these limitations with an \textit{allocentric} belief network, a mechanism that infers team state from the behavior of a witness which acts as an external reference shared by all agents. Allocentric observation is a popular method in both computer vision and robotics that entails representing observation from a centralized reference point \cite{8578982,zhao2021learning,doi:10.1126/scirobotics.adz7130,salahshour2025allocentric}. In this work, each agent applies the belief network to observations of the same external actor, providing a common reference for state estimation. This is in contrast to \textit{egocentric} methods where each agent observes different environment variables as needed -- representing the state of the art discussed in the previous passage. The difference between the two is illustrated in Figure \ref{fig:master_intro}. Inferring full state from allocentric observation offers two major improvements to the current state of the art: both divergence between agent estimates and complexity of mapping partial-observation to global state is reduced by inferring state from the same information. Our work will highlight that the proposed approach not only produces more agreeable state estimates among decentralized agents, but can more accurately predict general team state than methods that allow for observation of other team members, thereby enabling better coordination. We then leverage the lessons learned on smaller scale collaborative teams to demonstrate allocentric belief's ability to scale to team sizes where precise coordination is generally believed infeasible. 

\section{Results}\label{sec:results}


As established, decentralized coordination for communications denied environments can be greatly improved by incorporating a mechanism for estimating global state through partial observation. Many works have explored how to optimize this problem, however, no method is immune to diverging global estimates and degraded coordination. In this section we will demonstrate the ability of allocentric belief networks to outperform conventional model-based estimation by leveraging a centralized witness as the sole information channel. Building on this, we then demonstrate the network's ability to scale to team sizes much larger than what is currently feasible with standard methods.

We verify our method's ability to outperform conventional egocentric model-based approaches by testing it on five multi-agent environments. For each environment the desired collaborative behavior requires precise maneuvering of each team agent where even a single deviating agent can cause the system to fail. This requires group members to keep accurate global estimates in order to enable highly synchronized behavior. However, agents are only capable of observing a single actor at any point in time, and cannot communicate. Each of the environments were inspired by problems that are of interest to a range of different academic communities, and offer a number of variety of environmental challenges, they include predatory prey \cite{3295222.3295385}, drone caravan protection \cite{9615724,egerstedt2015coordinated}, football rondo circle \cite{kurach2020google}, firefighting management \cite{8593539,drones9010017}, and resource constrained market coordination \cite{arrow1951optimal, lee1997information, oroojlooyjadid2022deep}. These environments are discussed in greater detail in Section \ref{sec:methods}, and the Supplementary material.

In order to implement the allocentric belief network coupled with a decentralized RL control policy we propose a multi-agent RL training regime that exists in 2 key stages. First, we train a decentralized RL policy for the noisy fully-observable environment. This is done by employing imitation learning techniques that leverage expert knowledge in the form of a centralized, fully-observable controller in order to ensure a successful decentralized policy can be learned. Second, we deploy this policy in simulation, collect data on interactions between the team and the witness, and train the allocentric belief network on these data points. At execution each agent runs its own copy of the belief network and policy -- estimating the team’s state from observations of the witness alone -- removing any need for inter-agent communication. As a baseline, we implement in each environment an egocentric model-based estimator popular for decentralized, communications-limited systems \cite{10333311,ito2026robottask,krogius2021decentralized} in which unobserved teammates are propagated with the centralized expert $\pi_c$.  During testing of both the allocentric and egocentric methods the same implementation architecture is used on each decentralized agent. In the following subsections we will discuss the results from deployment of these models, and analysis of their performance. 

\subsection{Allocentric Belief Model Performance}

The first metric analyzed was overall success rates of each environment. Success is defined by either completing the desired task before episode cut off (predator-prey, and fire management environment), or avoiding early episode termination (caravan defense, football environments, and market coordination). Overall, the allocentric belief model massively out-performed the model-based state-estimation method and achieved overall success rates much closer to that of the fully-observable environment (Table \ref{tab:success}). Generally, teams using the model-based system will see coordination slowly deteriorate until the team reaches an unrecoverable state in which the episode ends early, or the state becomes too different from the data the model was trained on causing it to act unpredictably. This can be observed in the Supplementary material, where we provide several comparisons of the centralized controller, decentralized policy, allocentric, and egocentric methods on a number of randomly generated seeds. It is worth noting in the football environment that success rates for both the model-based baseline and the allocentric belief network drop substantially relative to the fully observable setting. Football has the most challenging dynamics of the five environments largely due to when the game engine prohibits ball passing. One possible explanation for the lower success rate in football is that the attacker responds primarily to the ball, making its trajectory less informative about the receiving players’ positions.

\begin{table}[h]
\caption{\textbf{Episode success rate (\%).} Brackets, 95\% CI; $n = 1{,}000$ episodes per method.}\label{tab:success}
\footnotesize
\setlength{\tabcolsep}{4pt}
\begin{tabular}{@{}lccc@{}}
\toprule
 & Alloc. & Egoc. & Full obs. \\ 
\midrule
Predator--prey & 92.0 {\tiny[90.2, 93.5]} & 13.3 {\tiny[11.3, 15.5]} & 100.0 {\tiny[99.6, 100.0]} \\ 

Football       & 36.8 {\tiny[33.9, 39.8]} & 0.0 {\tiny[0.0, 0.4]}    & 67.3 {\tiny[64.3, 70.1]} \\  

Drones         & 74.7 {\tiny[71.9, 77.3]} & 9.8 {\tiny[8.1, 11.8]}   & 99.1 {\tiny[98.3, 99.5]} \\  

Fire           & 71.0 {\tiny[68.1, 73.7]} & 49.3 {\tiny[46.2, 52.4]} & 98.6 {\tiny[97.7, 99.2]} \\ 

Market         & 97.5 {\tiny[96.3, 98.3]} & 38.3 {\tiny[35.3, 41.4]} & 97.5 {\tiny[96.3, 98.3]} \\  
\botrule
\end{tabular}
\end{table}

\begin{figure}[h!]
\centering
\includegraphics[width=0.98\textwidth]{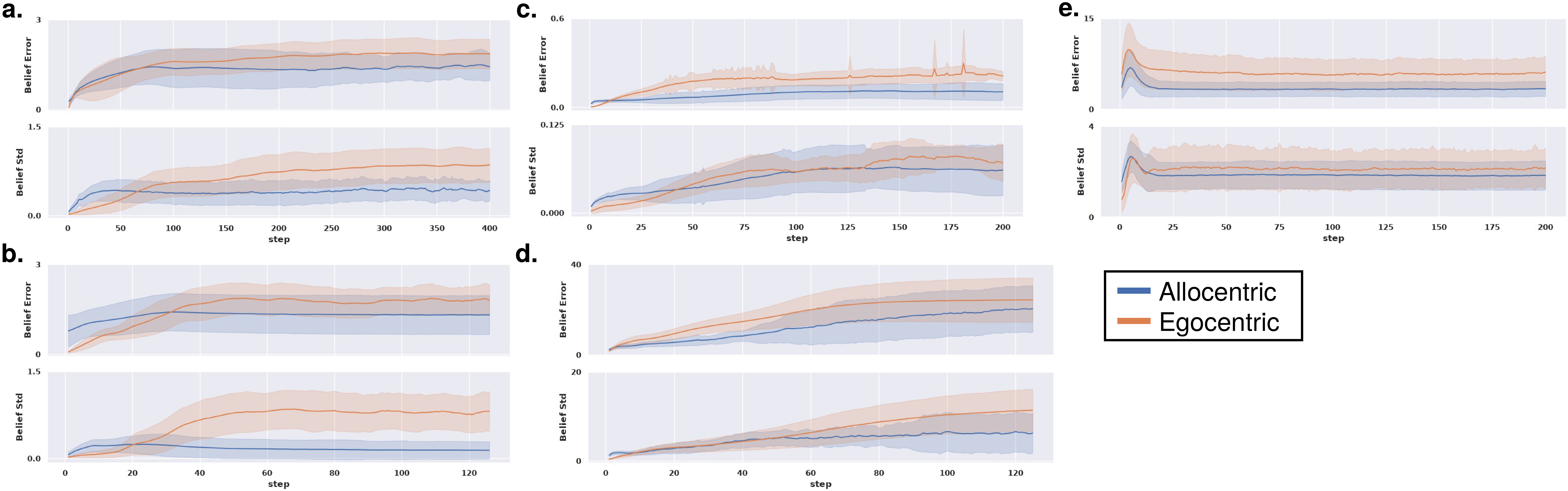}
\caption{The error and standard deviation of team state estimates, along with decomposed reward metrics for all environments. Shaded regions represent one standard deviation. Top-down left-right: \textbf{a)} predator-prey, \textbf{b)} drone caravan, \textbf{c)} football, \textbf{d)} fire fighting management, \textbf{e)} market coordination.}\label{fig:errors}
\end{figure}

In addition to success rate Figure \ref{fig:errors} provides a useful illustration demonstrating the consensus between agent world estimates when leveraging allocentric versus egocentric belief models. Here, belief error is the average sum of euclidean distance between the estimated team positions and true team positions. And, the belief standard deviation is the standard deviation for all agent estimates of the same agent at a given point in time. As Figure \ref{fig:errors} shows, for all five environments the error and standard deviation of estimates follow the same trend of being lower when deploying allocentric observation. The lower cross-agent dispersion indicates that the allocentric models produce more consistent estimates, alongside the reductions in estimation error shown in Figure \ref{fig:errors}. 

While the belief error and standard deviation give an informative picture of how the allocentric belief networks behave, and why they outperform egocentric observation both metrics consider only the estimates of the collaborative team. In practice, the model-based method must also predict the adversary's state. And, because witness and team behavior are coupled, as one estimate diverges this can cause a cascading error effect for all agent estimates. Observing the adversary directly removes the need to propagate an unobserved estimate of its state and provides measurements that may also contain information about the team. This feedback loop is a further limitation of model-based estimation in multi-agent environments; with allocentric belief, however, it can be easily mitigated by ensuring that one side of the loop (either the witness or the team) is always observed. Moreover, the decentralized policy may rely on statistical dependencies between state variables that the computed error and variance do not capture. In order to better understand the effects these factors have on model performance we must perform a more in-depth analysis on the estimates being provided and how they truly replicate the environment ground truth.

\subsection{Canonical Correlation Analysis of Task Dimensions}

In order to obtain a richer analysis on the key data maintained in the estimates provided by the allocentric and egocentric belief networks vs. the ground truth we perform a Canonical Correlation Analysis (CCA) on the data collected during implementation. The CCA provides us with an essential data point that allow us to better understand the gap in performance between the two state estimation methods: the key task directions $\boldsymbol{\phi}$ which are uncorrelated projections of the ground-truth state that are most strongly correlated with the decentralized fully-observable model's target actions. By comparing the fit of the allocentric vs. egocentric belief estimates projected onto each of the key directions we can analyze how well each preserves key information needed for multi-agent coordination.

\begin{figure}[h!]
\centering
\includegraphics[width=0.98\textwidth]{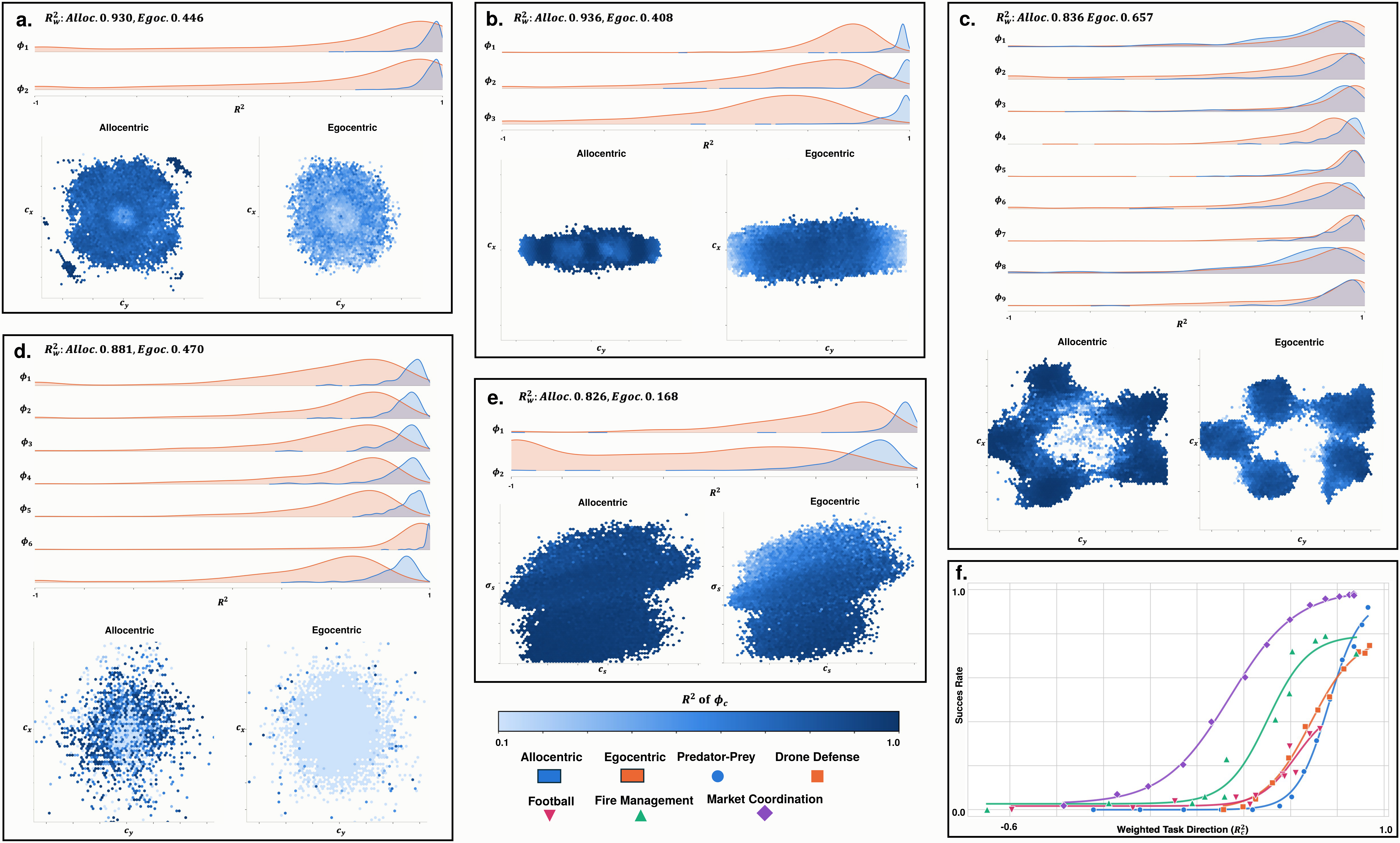}
\caption{Results from the Canonical Correlation Analysis (CCA), including ridgeline plots of state estimation methods fit to task directions, and heat maps of estimation data fit to team centroid direction $\phi_c$ over centroid x, y coordinate ($c_x$, and $c_y$). Data was collected over 1000 episodes. Top-down: \textbf{a)} predator-prey, \textbf{b)} drone caravan, \textbf{c)} football (only first 100 timesteps of each episode analyzed so that methods are compared on similar game stages), \textbf{d)} fire management, and \textbf{e)} market coordination. \textbf{f)} Results from our noise sweep, showing that coordination degrades with loss of task direction information in estimated state.}\label{fig:cca}
\end{figure}

Figure \ref{fig:cca} shows the results from this CCA analysis. In the figure for each environment we provide a ridgeline plot showing the $R^2$ value of how well the state estimates matched the ground truth data when projected into the direction for each of the tasks. For each environment the weighted $R^2$ average for both methods is provided as well. Looking at the figure it is clear that the allocentric methods maintain the task directions considerably better than egocentric. In addition to this, we analyze the performance of our method over increasing levels of noise injected into the partial observation. These results (Figure \ref{fig:cca}.f) show that as the information provided in the task directions is lost coordination does in fact degrade. 

Upon further analysis of the individual task directions, we found that most decomposed into 3 qualitative modes of information: the adversary, the team centroid, and the difference between individual agent states. Each of these can be represented as unique projections, and the same analysis of belief state fit to true state can be performed. Surprisingly, our results show that both methods perform equally poorly at predicting differences between individual agent states, however, allocentric consistently outperforms egocentric methods in predicting the team centroid. In addition, the allocentric methods unsurprisingly provides more accurate observation of the adversary. This is illustrated in Figure \ref{fig:cca}: for each environment, it shows the local $R^2$ of the centroid projection $\phi_c$ as a function of the true centroid position (x, y), over all agent-steps collected during evaluation. These results indicate that the allocentric models recover collective features more effectively than differences between individual agent states. In particular, improved centroid estimation accompanies weak recovery of inter-agent differences in both methods. This suggests that the performance gains depend on preserving task-relevant collective structure rather than accurately reconstructing every team member’s state.

\subsection{Scaling for Swarm Applications}

The knowledge gained from the CCA in the previous section suggests that our allocentric method -- while demonstrated so far on small collaborative teams -- should extend to any team whose state can be decomposed into a set of shared team-level variables, such as the centroid. This motivates representing larger teams through a compact set of collective variables, allowing us to evaluate coordination without estimating every team member’s state separately.

We evaluated this representation in a predator–prey swarm of 40 agents, and a market coordination environment with 60 producers. The training workflow for the decentralized policy and the concurrent belief network was unchanged, as was the objectives in each environment. However, the team is more than ten times larger. This makes coordination considerably more difficult, and egocentric methods become infeasible due to the sheer number of observations required. Rather than representing teams through individual agent states we describe them with team-wide statistics similar to the centroid. For the predator-prey swarm environment we describe the team as an ellipse with centroid $(c_x, c_y)$, orientation angle $\varphi$, and elongation parameters $e_1, e_2$. Similarly, for the scaled market coordination game we defined the team state by stockpile average, as well as its $10^{th}$ and $90^{th}$ percentile. Results for the full system are shown in Figure \ref{fig:scaled_results}.

\begin{figure}[H]
\centering
\includegraphics[width=0.98\textwidth]{scaled_results.png}
\caption{Results from the scaled predator-prey and market coordination environment. \textbf{a)} From top left going counter clockwise: an illustration of how swarm shape is described through an ellipsoid (top left), decomposed reward (top right), fit to centroid $\mu$ and orientation angle $\varphi$ over centroid ($c_x,c_y$), and elongation $e_1, e_2$ respectively (bottom right), and general success metrics over 1,000 runs (bottom left). \textbf{b)} From top left going counter clockwise: an illustration of how the distribution of producer stockpiles is represented by mean, $10^{th}$, and $90^{th}$ percentile (top left), general success metrics over 1,000 runs (top right), fit to average team stockpile $\mu_s$ (bottom right), and average agent decomposed reward over 1,000 runs (bottom left).}\label{fig:scaled_results}
\end{figure}

Upon analyzing the performance of the two scaled environments the strongest data point indicating its success is the near zero drop in success rate from the fully observable decentralized model. Furthermore, the decomposed reward plot in the top right of Figure \ref{fig:scaled_results}.a shows that on average individual swarm members contribute to the team objective as effectively as the centralized expert controller. 
Finally, we provide heat maps that allow us to visualize how well the centroid and orientation angle of the ellipsoid is predicted over several different ellipse parameters similar to the plots provided in Figure \ref{fig:cca}. When comparing the heat maps in the two figures we can see that we achieve similar $R^2$ consistency over the majority of centroid and elongation parameters. 

The results in these experiments show a quite remarkable trend: that even as team size grows -- which typically brings a sharp rise in environment complexity and a corresponding drop in performance -- collective behavior via allocentric belief transfers almost one-to-one despite the complete removal of inter-agent communication, and severe limitations on observation.

\section{Discussion}\label{sec:discussion}

Across five simulated environments, learned allocentric belief models improved decentralized coordination relative to the egocentric model-based baseline. The state-estimation analysis indicates that these gains accompany stronger recovery of collective features, particularly the team centroid, despite weak recovery of differences between individual agents. Experiments with 40-agent predator–prey swarms and 60-producer markets further demonstrate the use of compact collective representations for larger teams.

The observed actors follow fixed policies throughout training and evaluation. The present results therefore establish performance within those policy settings. Evaluating the estimators against actors that change their behavior, including actors that adapt to mislead the team, is an important direction for future work.
\section{Methods}
\label{sec:methods}

As discussed in previous sections of our work our proposed method works in two stages: first, a decentralized RL model $\pi^*$ is trained to provide collaborative control in a fully-observable environment. The model is then leveraged to train a belief model $b_{\phi}$ that takes partial observations of the witness, and predicts full team state. During implementation, each agent observes the witness, predicts team state through $b_{\phi}$, and uses this as observation to compute an appropriate action through $\pi^*$. In the following sections we will discuss in depth our methodology for both stages.

\subsection{Decentralized Collaborative MARL for the Adversarial Environment}

Decentralized collaborative control in communications-denied environments is often formalized as a Decentralized Partially Observable Markov Decision Process (Dec-POMDP) \cite{nayyar2014common,oliehoek2016concise}. In comparison to the far more studied centralized, fully observable multi-agent problem, which is often modeled as a single Markov Decision Process (MDP) and solved in polynomial time \cite{papadimitriou1987complexity}, Dec-POMDPs are NEXP-complete \cite{bernstein2002complexity}, and are much more difficult to solve. In our collaborative multi-agent Dec-POMDP witness environment, $N$ agents must collaboratively navigate a reactive, noncooperative environment while being restricted to observing a single environment agent at each timestep. The witness is assumed to follow a stationary policy $\pi_w$, and thus can be treated as hidden variables in the environment dynamics. The Dec-POMDP Bellman equation then becomes:

\begin{multline}
Q_i^*(o_i,a_i|\boldsymbol{\pi}_{-i}) = \sum_{s\in \mathcal{S}} p(s|o_i) \sum_{\boldsymbol{a}_{-i}\in\mathcal{A}^{N-1}}\boldsymbol{\pi}_{-i}(\boldsymbol{a}_{-i}|s) \\ \left[r(s,a_i,\boldsymbol{a}_{-i})+\gamma\sum_{s'\in \mathcal{S}}p(s'|s,a_i,\boldsymbol{a}_{-i})\max_{a_i'}Q_i^*(s,a'_i)\right]
\label{eq:marlbellman}
\end{multline}

where $Q^*$ is the expected utility of taking action $a_i$ given partial observation $o_i$, $r$ is a reward function, $\gamma$ is a discount factor, and all other agents policies are $\boldsymbol{\pi}_{-i}$ \cite{bellman1966dynamic}. Thus, in order to optimize for $Q^{*}$ a policy must learn two environment functions: the reward (which is given), and the transition function $p(s'|s_t,a_t)$ which must be implicitly learned through experience and can be decomposed to:

\begin{align}
    p(s'|s,a_i,\boldsymbol{a}_{-i}) = \sum_{\boldsymbol{a}_{-i}\in\mathcal{A}^{N-1}}p(s'|s,a_i,\boldsymbol{a}_{-i})\prod_{j\neq i}\pi_i(a_{-i}|s)
    \label{eq:long prob}
\end{align}

Unfortunately, Equation \ref{eq:long prob} and \ref{eq:marlbellman} are intractable for several reasons: first, agent policies are non-stationary which means $\boldsymbol{\pi}_{-i}$ may not be known or very difficult to predict throughout training, and second $p(s|o_i)$ which is also dependent on agent policies is unknown and cannot be learned during training. 
The combined effect is that the transition function, and belief probability cannot be learned, making it nearly impossible to learn a successful decentralized policy. 

The outlined problems cannot be overcome entirely, however, the equation can be simplified in order to make learning a decentralized policy more manageable. In order to do this we propose two environmental assumptions that greatly reduce the complexity of Equation \ref{eq:marlbellman}. First, we assume all collaborative agents follow the same decentralized greedy policy $\pi$. Second, we make a similar assumption that there exists a deterministic belief distribution $b_\phi$ that replaces $p(s|o_i)$. The simplified equation then becomes: 

\begin{align}
Q_i^*(o_i,a_i|\boldsymbol{\pi}_{-i}) = r(b_\phi(o_i),a_i,\boldsymbol{a}_{-i})+\gamma\sum_{s'\in \mathcal{S}}p(s'|b_\phi(o_i),a_i,\boldsymbol{a}_{-i})\max_{a_i'}Q_i^*(b_\phi(o_i),a'_i)
\label{eq:marlbellman_simp}
\end{align}

this formulation provides a significantly less complex version of Equation \ref{eq:marlbellman}. However, it is still not an easy problem to learn using standard RL training algorithms which assume the environment can be modeled as an MDP, where the environment is fully observable. Additionally, the joint reward function $r(b_\phi(o_i),a_i,\boldsymbol{a}_{-i})$ and transition probability $p(s'|b_\phi(o_i),a_i,\boldsymbol{a}_{-i})$ are complex non-stationary equations that rely on a belief estimate $b_\phi(o_i)$ which can complicate converging on an optimal policy. 

In order to guarantee that we are able to converge upon an optimal decentralized policy that demonstrates a desired set of behaviors for each environment we implement two crutches into our training environment. First, we train the initial model in a fully observable environment where agents receive the true state $s$, and second, we introduce a centralized controller $\pi_c$ that provides expert knowledge to the policy during training. Each controller is a heuristic function that takes the full state of the environment, and computes a set of $N$ desired team actions:

\begin{align}
\pi_c(s) = [a_{\pi_c,0},...,a_{\pi_c,N-1}] 
\label{eq:controller}
\end{align}

A more detailed description of the controllers for each environment can be found in the Supplementary material.

In order to leverage the expert information provided through the centralized controller we deploy two behavior cloning techniques in order to stabilize the environment by reducing non-stationarity, and keeping each decentralized agent's behavior within a bounded range of the centralized controller's. First, we pre-train the policy on noise-free data collected from implementing the centralized controller. Second, we implement a behavior cloning reward term in each agents joint reward:

\begin{align}
r_{BC} = w_{BC} \sum_{i\in N}[a_{\pi_c,i}=a_{\pi^*,i}]
\end{align}

where $[a_i^c=a_i^{\pi^*}]$ returns 1 if the controller action $a_i^c$ and policy action $a_i^{\pi^*}$ for agent $i$ are the same, and $w_{BC}$ is a static weight.

By implementing these assumptions and behavior cloning mechanisms we are successfully able to transfer desired collaborative control into a decentralized RL policy by using a standard PPO training algorithm.

\subsection{Allocentric Belief Model}

The decentralized policies that are trained using the method in the previous section assume a fully-observable environment, and receive full state $s$. In order to adapt these policies to a partially-observable environment as represented in Equation \ref{eq:marlbellman_simp} we must now train a deterministic belief network that can provide an estimate of $s$ from partial observations $o_i$. As previously stated, we do this through allocentric belief that only relies on observation of the witness. And, by ensuring largely similar observations are fed to identical belief networks, decentralized agents will possess a congruent world estimate which enables precise coordination. The models implemented are multi-layer perceptrons (MLPs) trained on experiences collected from testing the decentralized fully-observable policies. The belief network estimates team state from a history of witness observations and the known goal. In this section we define our generalizable state definition, and how our belief model is trained.

The full state of our environment is defined as the concatenation of all team member states $\boldsymbol{p}_i$, the witness' state $\boldsymbol{p}_w$, and any additional environment relevant information $\boldsymbol{p_e}$:

\begin{align}
\boldsymbol{s} = [\boldsymbol{p}_0,...,\boldsymbol{p}_{N-1},\boldsymbol{p}_w,\boldsymbol{p}_e]
\label{eq:state}
\end{align}

the action space is then described as a discrete set of directional vectors $\boldsymbol{a}$. As previously mentioned, in the communications denied environment we assume that each agent only has the resources to observe a single agent at a time. Our allocentric belief network is designed to predict the team state by only observing the adversary. Thus, we describe our observation space as:

\begin{align}
\boldsymbol{o}_{i,t} = [\boldsymbol{p}_{w,t-M}^i,...,\boldsymbol{p}_{w,t}^i,\boldsymbol{p}_e^i]
\label{eq:observation}    
\end{align}

where the superscript $i$ implies in relation to agent $i$, and $\boldsymbol{p}_{w,t-M}^i,...,\boldsymbol{p}_{w,t}^i$ is a time series of past observed witness states. Our allocentric belief network, must then provide an estimation of the missing state $\{\boldsymbol{p}_{j}^i|j=0,...,N-1,j\neq i\}$ denoted as $\hat{\boldsymbol{y}}_{t}$. 

Our belief model is a fully connected neural network that takes observation $\boldsymbol{o}_{i,t}$ as input and returns the missing team state as an array $\boldsymbol{b}_{i,t}$. In order to provide temporal stability among estimates, we also append the belief models previous estimate of the team state $\boldsymbol{b}_{i,t-1}$ to the input. Network updates are done using a permutation invariant loss function:

\begin{align}
\mathcal{L} = \min_{\sigma \in S_N}\sum_j^{N-1}\mathcal{L}_2(\boldsymbol{b}_{i,t}(j),\hat{\boldsymbol{y}}(j))
\label{eq:loss}
\end{align}

where $\boldsymbol{b}_{i,t}(j)$ is agent $i$'s belief of agent $j$, $\hat{\boldsymbol{y}}$ is the ground truth, $\sigma$ represents a possible permutation of missing agent positions, $S_N$ is all potential permutations of team agents, and $\mathcal{L}_2$ is the L2 loss for an individual agents state. For environments where agents are not assigned to unique roles, and reward is purely based off of group performance we find a permutation invariant loss significantly improves prediction accuracy. Thus, an agents potential position is interchangeable with any other agents to receive the same observation.

During data collection to train the model, experiences from several thousand episodes using $\pi^*_{dec}$ are collected. This process is fairly straightforward: data in the format of $[\boldsymbol{o}_i,\boldsymbol{b}_{i,t-1}]$ is collected as input, and the missing state $\{\boldsymbol{p}_{j}^i|j=0,...,N-1,j\neq i\}$ is collected as data labels. However, there is one caveat due to the fact that the belief estimate $\boldsymbol{b}_{i,t-1}$ cannot be generated before the belief network is trained. In order to circumvent this the training data input collected is of the form:

\begin{align}
\boldsymbol{x}_t = [\boldsymbol{p}_{w,t-M}^i,...,\boldsymbol{p}_{w,t}^i,\boldsymbol{p}_e^i, \boldsymbol{p}_{j}^i+ \mathcal{N}(\mu,\sigma)]
\label{eq:belief_in}
\end{align}

where $\boldsymbol{p}_{j}^i+ \mathcal{N}(\mu,\sigma)$ represents the true team state plus some gaussian noise. This allows the belief model to better generalize when taking its own noisy predictions as input. Additionally, during training the model has 2 prediction heads, one that uses $\boldsymbol{p}_{j,t}$ as a label, and a second that uses $\boldsymbol{p}_{j,t-1}$. This encourages the model to implicitly include predicted team actions into its training predictions, and improves overall performance of the model.

During implementation, model performance is then evaluated using several metrics including belief error:

\begin{align}
\epsilon_b = 
\frac{1}{N}\sum_{i=N}\sum_{j\neq i} \frac{||\boldsymbol{s}_j-\boldsymbol{b}_{i,j,t}||}{N-1}
\end{align}

where $\boldsymbol{s}_j$ is agent $j$'s true state, and $\boldsymbol{b}_{i,j,t}$ is agent $i$'s estimated state of agent $j$. Belief standard deviation:

\begin{align}
\sigma_b = 
\sqrt{\frac{\sum_{i=N}\sum_{j\neq i}||\boldsymbol{\mu}_{i,t}-\boldsymbol{b}_{j,i,t}||^2}{N(N-1)}}
\end{align}

where $\boldsymbol{\mu}_{i,t}$ is the average estimate of agent $i$'s state among all other agents. And, a decomposed reward metric:

\begin{align}
\kappa_i = r_i|\underset{\boldsymbol{s}_j \in \mathcal{S} | j\neq i}{\operatorname{argmin}} d(\boldsymbol{s}_i,\boldsymbol{s}_j)||
\end{align}

where $r_i$ is a decomposed reward that measures the effect on team performance contributed by agent $i$, and the argmin term represents the shortest distance between agent $i$'s state and another team member. This is added in order to ensure agents showing near identical behaviors score poorly.

For further analysis of model performance we identify task-relevant directions of the state with canonical correlation analysis (CCA) on the fully-observable rollouts. 
In order to do this, we compute the CCA between actions chosen by $\pi^*$ and $\boldsymbol{s}_t$ and keep the canonical directions $\boldsymbol{f}_k$ with squared canonical correlation $\sigma_k^2 > 0.1$. These directions are fixed and reused for each future analysis. To score an estimator, we project its state estimate $\boldsymbol{b}_t$ and the true state $\boldsymbol{s}_t$ onto each direction $\boldsymbol{f}_k$, and compute the $R^2$ value:

\begin{align}
R^2_k = 1- \frac{\mathbb{E}[((\boldsymbol{b}_t-\boldsymbol{s}_t)\cdot \boldsymbol{f}_k)^2]}{\mathrm{Var}(\boldsymbol{s}_t\cdot \boldsymbol{f}_k)}
\label{eq:rsq}
\end{align}

This measures how accurately the belief reproduces the state along the directions the policy's actions depend on. A coverage score, which tells us how well the estimates fit over all task directions can be computed by taking the weighted average using the squared correlation values:

\begin{align}
R^2_w = \frac{1}{\sum_k \sigma^2_k}\sum_k \sigma^2_k R^2_k
\label{eq:wrsq}
\end{align}

Finally, we can analyze other environment parameters such as centroid position using Equation \ref{eq:rsq}, the directions $\boldsymbol{f}_c$ for which can be found by rewriting the equation:

\begin{align}
c_x = \frac{\sum_N x_i}{N}
\end{align}

as a linear mapping in the form of $\boldsymbol{c}=\boldsymbol{C}\boldsymbol{s}$ where normalized rows of $\boldsymbol{C}$ give one direction per spatial dimension. Similarly, this can also be done for adversary position and differences between individual team members.

\subsection{Training Environments}

As mentioned in Section \ref{sec:results} we implement our method over several different environments taken from numerous fields from robotics to economics in order to convey allocentric belief's potential to impact a number of different scenarios. Extended Data Fig. 1 includes an overlay from a single episode in each environment. The five scenarios included are:

\begin{enumerate}
    \item \textbf{predator-prey}: predator-prey environments are a popular benchmark in the Multi-Agent Reinforcement Learning (MARL) community \cite{3295222.3295385}. We implement a variant of the predator-prey environment where a team of agents must collaboratively corral an adversary towards a predetermined goal point. This environment is implemented on the Farama Foundation's mpe2 environment framework \cite{mpe2}.
    \item \textbf{Drone Caravan Protection}: a notable adversarial environment that requires multi-agent collaboration \cite{9615724,egerstedt2015coordinated}. This environment entails a team of drones that must successfully create a barrier to protect an array of target areas from an adversarial drone. This environment offers both more challenging physical dynamics, but also adversarial dynamics which are reactive not just to the teams behavior but also the environment state. The environment is implemented on the pybullet-drones library which simulates the 4DOF dynamics of a quadcopter drone \cite{panerati2021learningflygym}.
    \item \textbf{Rondo Circle}: In this environment we have recreated a rondo circle -- a classic soccer training drill in which team members arranged in a circle must pass the ball to other members without being stopped by a defender who starts in the center of the circle. The team consists of five collaborative members, and coordination is highly dependent on environment variables such as ball position. This environment is implemented using Google's research-football engine \cite{kurach2020google}.
    \item \textbf{Fire Fighting Management}: Optimal control for drone based fire fighting management has been a topic of interest to the robotics community for its partial observability, challenging environment dynamics, and need for well coordinated behavior \cite{8593539,drones9010017}. In this work, we model the environment based off of the literature cited which treats the fire as a randomly spreading body within a grid world. Drones must evenly distribute themselves across the fire in order to extinguish outer pixels and prevent further spreading. To complicate the environment further drones are only capable of observing agents or the fire within a specific range of their location, making the environment even less observable than other environments.
    \item \textbf{Resource Constrained Market Coordination}: In this environment we explore the resource constrained market problem where producers must use minimal knowledge of available resources and other producer stockpiles to collectively meet market demand \cite{arrow1951optimal, lee1997information, oroojlooyjadid2022deep}. In this scenario, a resource required in order to make a specific product is entirely controlled by a third party. Decentralized producers must then coordinate how much of the constrained resource they consume at each step in order to ensure that total product demand is met while all producers keep enough material to stay in business. The resource provider on the other hand is aware of all producers and their conversion efficiencies, and releases the minimum amount of resource at each step in order to meet market demand. The resource provider’s release decisions depend on the producers’ aggregate state, making those decisions a potential source of information about the other producers.
\end{enumerate}

In addition to these five environments, we demonstrate our methods ability to scale on both the predator-prey environment and market coordination environment.

\subsection{Egocentric Baseline}

As a baseline an egocentric model-based state-estimation method was also designed for each environment following the structured approaches of Catellani et al. and Ito et al. \cite{10333311,ito2026robottask}. The model-based method employs an egocentric observation scheme where each agent chooses which actor to observe at each timestep. Because communication and observation are restricted and each agent's egocentric observations arrive in an unstructured order, an MLP similar to the allocentric approach would be unable to predict the global state. Instead, every agent runs an independent recursive estimator that uses known models of the team and of the environment to carry its estimates forward between observations. All estimators share the same general structure, and only differ in how belief is propagated.

In each egocentric estimator teammates are assumed to follow the centralized expert controller $\pi_c$, and non-team components a known model $\pi_w$ which is the true policy of the witness. For each state agent $i$ reconstructs the input $\boldsymbol{s}_{t}$ that the agent's policy would receive, evaluates the corresponding model to obtain $a_{i,t}$ using $\pi_c$ or $\pi_w$, and propagates the belief through the transition model $f$:

\begin{equation}
\boldsymbol{b}_{i,t}^- = \int p_f\!\left(\boldsymbol{s}' \mid \boldsymbol{s}, a_{i,t}\right) \, \boldsymbol{b}_{i,t-1} \, d\boldsymbol{s}.
\label{eq:predict}
\end{equation}

where $\boldsymbol{b}^-_{i,t}(j)$ is the prior belief before observation. At each timestep, individual agents have the option to choose which agent to observe (witness or specific team member). This is done by choosing the agent that the observing agent is least confident in its estimation of. A confidence value $c_{i,t}(j)$ which represents agent $i$'s confidence in its estimation of agent $j$ at time $t$ is used for this. $c_{i,t}(j)$ is the inverse of a measure of the spread $\sigma$ of the marginal belief over $\boldsymbol{b}_{i,t}(j)$:

\begin{align}
c_{i,t}(j) = \frac{1}{\sigma\!\left(\boldsymbol{b}^-_{i,t}(j)\right)}
\end{align}

and is computed for every entity in the environment at each timestep. In order to prevent an agent from fixating on one entity, a random entity is selected when all confidences are tied or after the same entity has been observed for several consecutive timesteps.

When an agent receives partial observation $\boldsymbol{o}_{i,t}(j)$ of the selected entity the observing agent's belief is updated according to the measurement model of that entity following Bayes rule: 

\begin{equation}
\boldsymbol{b}_{i,t}(j) \propto p\!\left(\boldsymbol{o}_{i,t}(j) \mid \boldsymbol{s}_t, j\right) \boldsymbol{b}^-_{i,t}(j)
\label{eq:correct}
\end{equation}

where $p\!\left(\boldsymbol{o}_{i,t}(j) \mid \boldsymbol{s}_{j,t}\right)$ is a predetermined function that varies based on the environment being modeled and the environmental aspect $j$ being observed.

\subsection{Use of large language models}
Large language models (LLMs) were used to assist with debugging and refactoring the code used to run the experiments, and to refine the code used to generate figures. LLMs were not used to generate data, results or figure content.

\subsection{Statistical Reporting}

All reported results were computed from 1,000 randomly initialized episodes per environment and method (fully observable, allocentric and egocentric). Initial states were not matched across methods. Success rates are reported with 95\% Wilson score confidence intervals. The results presented were obtained from a single trained decentralized policy and belief network. However, results were repeated over 3 random seeds, all of which converged to successful policies, and similar performance metrics. Statistics therefore reflect variability across initial conditions, not across training runs. These results can be found along with the models trained in the github repository provided in our data availability disclaimer. 

\section{Supplementary Material}

In our Supplementary material we include several documents, videos, and the code base used to implement these experiments. For each environment trained we provide an appendix discussing the environment dynamics and how the team and witness interact. Additionally, for each experiment (5 base environments, and 2 scaled), we provide videos of the centralized controller, fully-observable decentralized RL, Allocentric, and Egocentric schemes over four different randomly initialized episodes. This is done so readers can qualitatively compare the difference in performance between the four methods when starting from identical states. For the scaled experiments, egocentric methods are not provided due to the impracticality of implementing them on a team with such size. In addition to this, we provide the code used to perform these experiments, as well as the data collected from implementation that was used to generate the results, and models trained. For reviewers who wish to retrain their own models, the documentation in the repository should provide sufficient instructions for using the proper bash scripts to do so. 

\newpage
\begin{appendices}
\label{sec:appendix}
\section{Predator Prey Environment Controllers}\label{secA1}

\begin{figure}[h]
\centering
\includegraphics[width=0.98\textwidth]{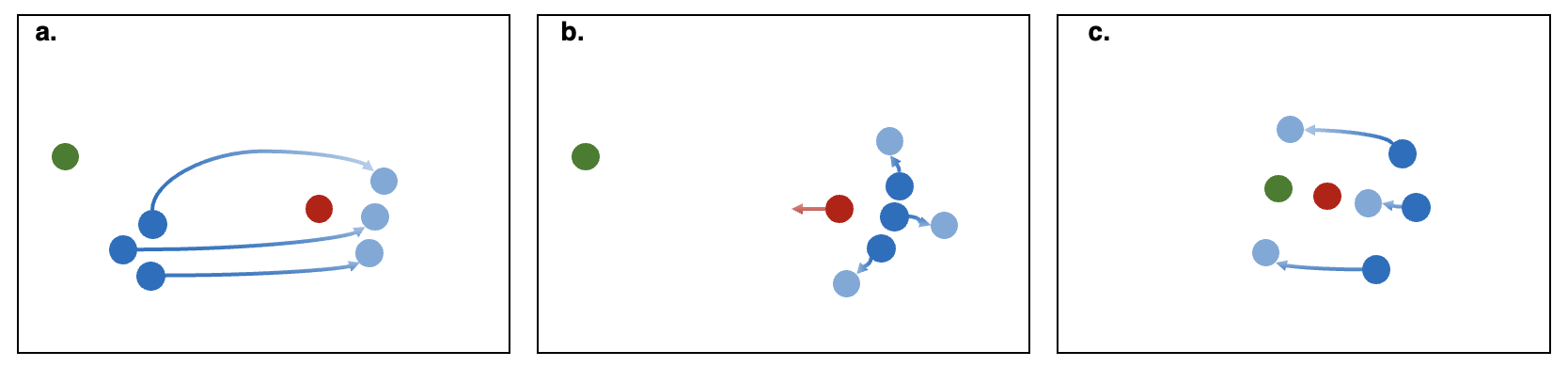}
\caption{An illustration of each of the 3 stages of our ``push and flank'' controller. \textbf{a)} represents the initial positioning. \textbf{b)} shows the approach of the target towards the goal. \textbf{c)} shows the final stage where agents surround the target to hold it at the goal.}\label{fig:predator_prey_control}
\end{figure}

The teams goal in the predator prey environment is to successfully "coral" the adversary towards a desired goal point, thus, the desired behavior captured by the controller is a "push and flank" strategy. In this behavior two agents move to either side of the adversary keep its movement horizontally stable, while a third agent pushes it from behind towards the desired goal point. In order to achieve this, there are 3 stages: initial positioning, the approach, and final surrounding of the goal. This is depicted in Figure \ref{fig:predator_prey_control}

For each stage, the desired agent target positions are computed using the same equation:

\begin{align}
\boldsymbol{p}_{t,i} = \boldsymbol{p}_a - c_{1,i}\hat{\boldsymbol{p}}_a^g + c_{2,i}\hat{\boldsymbol{p}}_{a,\perp}^g
\label{eq:predator_prey_team_control}
\end{align}

where $\boldsymbol{p}_a$ is the position of the adversary, $\hat{\boldsymbol{p}}_a^g$ is the unit vector from the adversary to the goal, $c$ is a constant, and $\boldsymbol{p}_i$ is the $i^{th}$ target position. Target positions are then assigned to agents by computing the closest position to each agent, no two agents can be assigned the same target.

Once a desired target position is chosen for each agent, actions can be computed from the discrete action space by computing the minimum angle between desired vector and discrete action:

\begin{align}
a_i = \underset{a \in \mathcal{A}}{\operatorname{argmin}}\; \cos^{-1}\left(\frac{\boldsymbol{p}_{t,i}^{p_i}\cdot a}{||\boldsymbol{p}_{t,i}^{p_i}||\,||a||}\right)
\label{eq:pick_discrete_action}
\end{align}

Conversely, the adversary's control is computed through a potential field vector algorithm, where each team member creates a negative force on the adversary and the sum of all forces provides a desired vector for the predator:

\begin{align}
    \boldsymbol{F} = \sum^N_{i=0} \boldsymbol{K}\frac{\boldsymbol{p}_{i}^a}{||\boldsymbol{p}_{i}^a||^\alpha}
\label{eq:predator_prey_adv_control}
\end{align}

where $\boldsymbol{K}$ is a gain, $\boldsymbol{p}_{i}^a$ is the position of the $i^{th}$ agent relative to the adversary, and $\alpha$ is a exponential constant.

\newpage

\section{Caravan Defense Environment Controllers}\label{secA2}

\begin{figure}[h]
\centering
\includegraphics[width=0.98\textwidth]{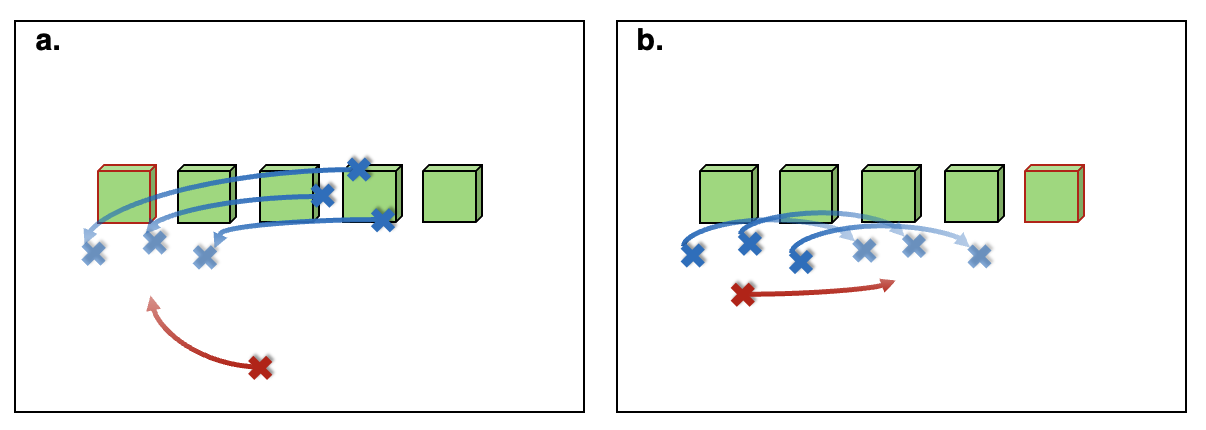}
\caption{An illustration of 2 of the key behaviors captured by the heuristic controllers employed in the caravan defense environment. \textbf{a)} represents the initial positioning where team agents randomly initialized around a point over one of the boxes must predict the adversary's movement and move in position to successfully block it from reaching a goal box. \textbf{b)} the environment state after the agents have successfully blocked the adversary's first attempt, and it must recompute an optimal goal box to approach forcing the team to again predict its movements and form a protective barrier}\label{fig:drone_control}
\end{figure}

In the drone caravan defense environment, the goal of the team is to successfully create a barrier between five distinct goal areas, and an adversarial drone. The adversarial drone is controlled by a potential vector field algorithm where it is attracted to the least defended area, while team members have a repulsive effect on the adversary. Thus, the desired strategy of the team is to for a barrier where agents are evenly distributed in a line between the goal area the adversary is targeting and the adversary itself. This strategy is illustrated in Figure \ref{fig:drone_control}.

The team controller is similar to the controller for the predator prey environment described in Equation \ref{eq:predator_prey_team_control}, however, adversary deviation from the center goal is clipped if it exceeds a euclidian distance of more than $d_{max}$: 

\begin{align}
\boldsymbol{p}_{t,i} = \hat{\boldsymbol{p}}_{t,i} min(||\boldsymbol{p}_{t,i}||,d_{max})
\label{eq:caravan_team_control}
\end{align}

where in this environment $\boldsymbol{p}_{t,i}$ is computed using Equation \ref{eq:predator_prey_team_control}, and $g$ represents the closest goal to the adversary's position. Positions are then matched to their corresponding team agent by sequentially choosing the closest position given team state, and a discretized action is chosen using Equation \ref{eq:pick_discrete_action}. Similarly, the adversary's control is based off of a potential vector field method  like Equation \ref{eq:predator_prey_adv_control} with an additional attraction term:

\begin{align}
    \boldsymbol{F} = \sum^N_{i=0} \boldsymbol{K}\frac{\boldsymbol{p}_{i}^a}{||\boldsymbol{p}_{i}^a||^2} + \boldsymbol{K}\boldsymbol{p}_g^a
\label{eq:caravan_adv_control}
\end{align}

where $\boldsymbol{p}_g^a$ is the position of the nearest goal to the adversary in relation to the adversary's position that is greater than a minimum distance from the closest team member.

\newpage

\section{Rondo Circle Environment Controllers}\label{secA3}

\begin{figure}[h]
\centering
\includegraphics[width=0.98\textwidth]{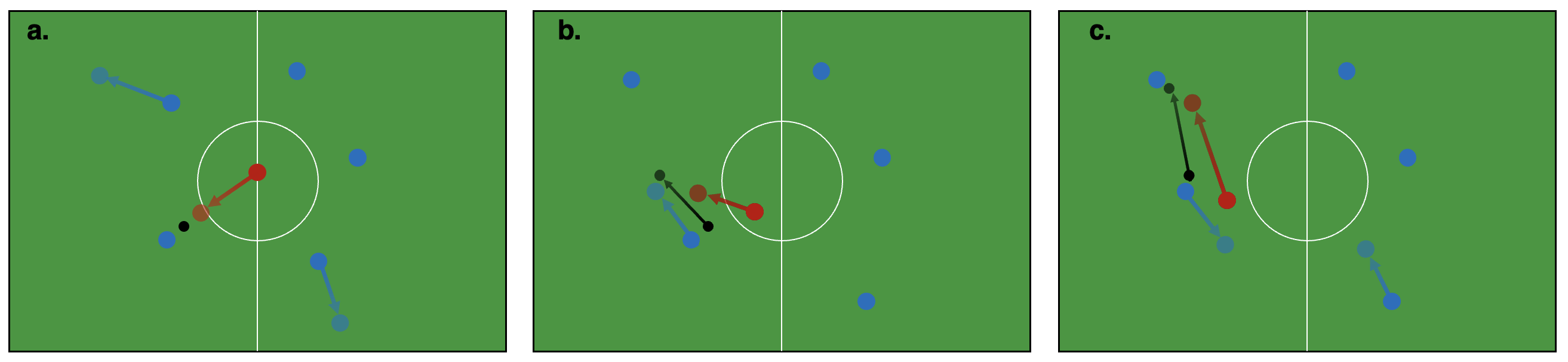}
\caption{An illustration of the 3 key behaviors captured by the football heuristic controller. \textbf{a)} Collaborative agents to the left and right of the ball owner will increase their distance from the center of the circle in order to open up more space for passing. \textbf{b)} The ball owner will pick an agent to move towards that has the greatest difference in angle between the two vectors leading from itself to the collaborative agent and itself to the adversary. \textbf{c)} The ball owner will pass the ball to the chosen agent once the angle between its heading and the selected agent is within a desired threshold. For all stages the adversary follows the ball and attempts to steal it once within a close enough range.}\label{fig:rondo_control}
\end{figure}

The rondo circle environment rules of play are taken from a classic football training drill where $n$ defenders are arranged in a circle with one attacker in the center. The defenders must then continue to pass the ball to other players on the team while the attacker attempts to steal the ball at any point in time. In order to emulate this behavior, we have broken the rondo game into three stages that are described in Figure \ref{fig:rondo_control}.

Each team agent is assigned a fixed anchor point evenly spaced around a circle centered in the middle of the field. At each time step the role of an agent is set by the index $i^*$ of the current ball owner. Agent $i^*$ owns the ball if the game engine reports that it is in possession, or if the ball is stationary and within a distance $d_{own}$ of the agent. Agents that do not own the ball move to a target position given by their anchor point:

\begin{align}
\boldsymbol{p}_{t,i} =
\begin{cases}
\lambda\, \boldsymbol{p}_{anc,i} & \text{if } (i^*-b) \bmod n \in \{1, n-1\} \\
\boldsymbol{p}_{anc,i} & \text{otherwise}
\end{cases}
\label{eq:rondo_team_control}
\end{align}

where $\lambda > 1$ is an expansion ratio, and $\boldsymbol{p}_{anc,i}$ is agent $i$'s anchor point. The two agents directly to the left and right of the ball owner therefore move outward, opening space for a pass (Figure \ref{fig:rondo_control}.a), while the remaining agents hold their positions on the circle.

The ball owner instead picks a receiving teammate based on how well the adversary covers each passing lane. The receiver $j^*$ is the teammate whose passing lane has the greatest angular separation from the adversary:

\begin{align}
j^* = \underset{j \in N \backslash i^*}{\operatorname{argmax}}\; \cos^{-1}\left(\frac{\boldsymbol{p}_j^{i^*}\cdot\boldsymbol{p}_a^{i^*}}{||\boldsymbol{p}_j^{i^*}||\,||\boldsymbol{p}_a^{i^*}||}\right)
\label{eq:rondo_receiver}
\end{align}

where $\boldsymbol{p}_j^{i}$ and $\boldsymbol{p}_a^{i}$ are the positions of teammate $j$ and the adversary relative to the ball owner agent $i$ (Figure \ref{fig:rondo_control}.b). The ball owner then moves in the direction of agent $j^*$ until the angle between its heading and the vector $\boldsymbol{p}_{j^*}^i$ is within a minimum passing threshold $\theta_{pass}$ (Figure \ref{fig:rondo_control}.c). 

For every movement command, a discrete action is chosen from the eight compass directions with Equation \ref{eq:pick_discrete_action}. If $||\boldsymbol{p}_{t,i}^{p_i}||$ falls below a small dead zone $\epsilon$ the no-op action is chosen instead so agents do not oscillate around their targets.

Unlike the previous two environments, the adversary is not controlled by a potential vector field. It pursues the ball directly, choosing its discrete action with Equation \ref{eq:pick_discrete_action} using the desired vector

\begin{align}
\boldsymbol{F} = \frac{\boldsymbol{p}_{b}^{a}}{||\boldsymbol{p}_{b}^{a}||}
\label{eq:rondo_adv_control}
\end{align}

where $\boldsymbol{p}_{b}^{a}$ is the position of the ball relative to the adversary. A steal is resolved by the game engine when the adversary reaches the ball, and the episode ends as soon as the adversary gains possession.

\newpage

\section{Fire Fighting Management Environment Controllers}\label{secA4}

\begin{figure}[h]
\centering
\includegraphics[width=0.9\textwidth]{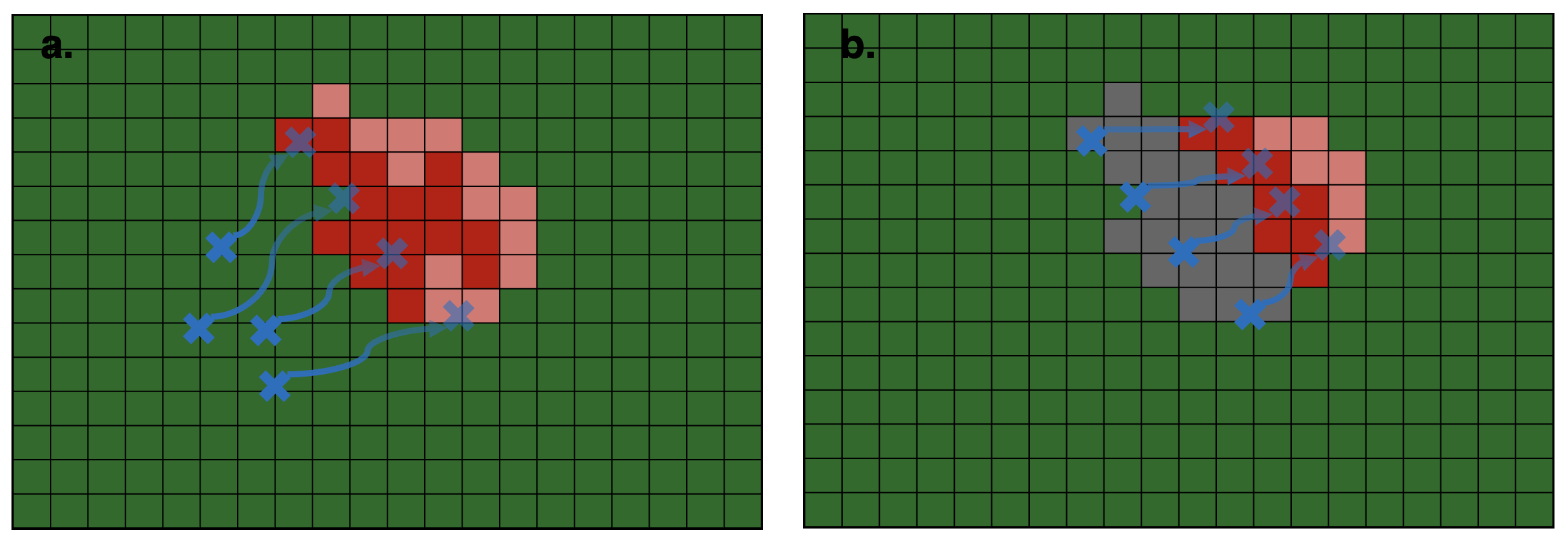}
\caption{An illustration of the 2 key behaviors captured by the fire management heuristic controller. \textbf{a.} Drones begin in an unorganized cluster, and arrange themselves in a line starting at the edge of the fire. \textbf{b.} The line of drones will sweep over the fire, deploying fire retardant when necessary, and putting out any remaining red pixels. Once pixels are extinguished, they remain black for the rest of the episode and cannot relight.}\label{fig:fire_control}
\end{figure}

The firefighting environment simulates a team of cooperative drones working together to contain an uncontrolled, spreading fire. The drones spread out into a single line that sweeps across the fire, deploying fire retardant where needed. The fire, in turn, spreads stochastically with a higher probability of spreading in the direction away from the drones, and a lower probability of spreading back towards them. This was implemented in order to simulate downwash effects from the drones propellers. Unique to this environment, agents are only capable of observing other agents or the environment within a certain proximity of themselves, making the environment even less observable than the others used in the work. Figure \ref{fig:fire_control} illustrates the two stage process performed by the centralized heuristic controller.

The environment is a grid with $n$ drones, and each cell is either unburnt (green), burning (red), or extinguished (black). Let $\mathcal{R}$, $\mathcal{G}$, and $\mathcal{B}$ be the sets of red, green, and black cells. The team controller places the drones on a single line perpendicular to the axis running from the drones' centroid $\boldsymbol{p}_d$ to the fire's centroid $\boldsymbol{p}_f$, the mean position of all cells in $\mathcal{R}$. Similar to Equation \ref{eq:predator_prey_team_control}, the ideal position of the $i^{th}$ slot on the line is:

\begin{align}
\tilde{\boldsymbol{p}}_{t,i} = \boldsymbol{p}_f + s\,\hat{\boldsymbol{p}}_d^f + o_i\,\hat{\boldsymbol{p}}_{d,\perp}^f, \qquad o_i = \left(i - \frac{n-1}{2}\right)\Delta + \bar{o}
\label{eq:fire_team_control}
\end{align}

where $\bar{o}$ is the mean lateral offset of the red cells near the line, and $\Delta$ is the spacing between slots. The line is placed at a depth $s$ into the fire shared by the whole team set to a low percentile of the red cells' positions along this axis so that the line uniformly advances through the fire. Each ideal slot is then snapped to an actual burning cell, preferring cells a short distance inside the fire over cells on its boundary:

\begin{align}
\boldsymbol{p}_{t,i} = \underset{r \in \mathcal{R}_{line}}{\operatorname{argmin}}\; \left|\boldsymbol{p}_r^{f}\cdot\hat{\boldsymbol{p}}_{d,\perp}^f - o_i\right| + \beta\left[h\left(d_{\mathcal{G}}(r)\right) + h\left(d_{\mathcal{B}}(r)\right)\right]
\label{eq:fire_slot_snap}
\end{align}

where $\mathcal{R}_{line}$ is the set of red cells within $d_{min}$ of the line that have not already been assigned to a slot, $d_{\mathcal{G}}(r)$ and $d_{\mathcal{B}}(r)$ are the distances from cell $r$ to the nearest green and black cells, and $\beta$ is a predetermined coefficient. The depth cost $h$ strongly penalizes cells closer than $d_{lo}$ to the boundary, and gradually penalizes cells deeper than $d_{hi}$:

\begin{align}
h(d) =
\begin{cases}
C + (d_{lo} - d) & \text{if } d < d_{lo} \\
\max(0,\, d - d_{hi}) & \text{otherwise}
\end{cases}
\label{eq:fire_depth_cost}
\end{align}

where $C$ is a large constant. Unlike the previous environments, slots are not assigned greedily. The assignment is the permutation that minimizes the total distance between drones and their slots.

A drone deploys fire retardant whenever a red cell lies within the extinguishing radius $r_{ext}$ of its position. Otherwise it moves towards its assigned slot. Once it is within $r_{ext}$ of the slot, it moves towards the nearest red cell instead (Figure \ref{fig:fire_control}.a). Because the environment is a grid, the discrete action is the element-wise sign of the desired vector, $\operatorname{sign}(\boldsymbol{p}_{t,i}^{p_i})$, which maps directly onto one of the eight neighboring cells. Each red cell within $r_{ext}$ of a drone that deploys retardant is extinguished with probability $e^{-\kappa d}$, where $d$ is its distance from the drone and $\kappa$ is a decay rate.

As mentioned, the fire follows a stochastic policy for spreading. At each time step, each green cell may be ignited independently by each of its eight neighbors that is burning. The probability that a red neighbor at offset $\boldsymbol{\delta}$ ignites green cell $g$ is biased so that the fire grows away from the team:

\begin{align}
P(g \mid \boldsymbol{\delta}) = \operatorname{clip}\left(p_s\left(1 + \gamma\,\frac{-\boldsymbol{\delta}}{||\boldsymbol{\delta}||}\cdot\hat{\boldsymbol{p}}_d^{g}\right),\, 0,\, 1\right)
\label{eq:fire_adv_control}
\end{align}

where $p_s$ is the base spread probability, $\gamma$ is a directional bias, $-\boldsymbol{\delta}$ is the direction of growth from the burning neighbor into cell $g$, and $\hat{\boldsymbol{p}}_d^{g}$ is the unit vector from the drones' centroid to $g$. Growth directly away from the team therefore happens with probability up to $(1+\gamma)p_s$, while growth directly towards the team is suppressed. 

\newpage

\section{Market Coordination Environment Controllers}\label{secE}
 
The team's goal in the market coordination environment is to keep every company's stockpile above a minimum level while together selling at least a fraction $\rho$ of market demand. Each company converts material into product at a different, hidden rate $k_i$, so the desired behavior captured by the controller is a ``fair share'' strategy. In this behavior each company orders only the share of the released material it needs to produce an equal part of demand, so that efficient companies do not starve inefficient ones, and the team sells slightly above the minimum demand fraction to protect its shrinking stockpile. In order to achieve this, the controller has 3 stages: ordering, setting a team sales target, and splitting sales across companies.
 
For each company, the order fraction is its share of the material the team needs, plus a small balancing term that keeps stockpiles together:
 
\begin{equation}
f_i = \frac{1/k_i}{\sum_{j=1}^{N} 1/k_j} + b\,\frac{\bar{S} - S_i}{M_t}
\label{eqE1}
\end{equation}
 
where $k_i$ is the conversion rate of the $i^{th}$ company, $S_i$ is its stockpile, $\bar{S}$ is the mean stockpile across the team, $M_t$ is the material released by the supplier, and $b$ is a gain. The team's sales target is then set below demand while the team stockpile is below a healthy level, but never below a safety margin $\delta$ above $\rho$:
 
\begin{equation}
P^{*}_t = \mathrm{clip}\!\left(D_t - \kappa\,\frac{S_H - S_{team}}{H},\; (\rho+\delta)D_t,\; D_t\right)
\label{eqE2}
\end{equation}
 
where $D_t$ is market demand, $S_H$ is the healthy team stockpile, $\kappa$ is a constant, and $H = \frac{1}{N}\sum_{j} 1/k_j$ is the average material needed per unit of product. Each company produces an equal share of this target, selling the fraction $u_i = P^{*}_t / (N k_i S'_i)$ of its stockpile after ordering, $S'_i$. Actions are then chosen from the discrete action space by selecting the closest allowed order and sell settings:
 
\begin{equation}
a_{t,i} = \operatorname*{argmin}_{(f,u)\in\mathcal{A}} \left( |f - f_i| + |u - u_i| \right)
\label{eqE3}
\end{equation}
 
Conversely, the adversary is a supplier that releases only the material the team needs to meet demand with an equal split, and withholds any team stockpile above a target level $S_{cut}$:
 
\begin{equation}
M_t = H D_t (1 + \epsilon_t) - \min\!\big(\eta \max(0,\, S_{team} - S_{cut}),\; c_{max} H D_t\big)
\label{eqE4}
\end{equation}
 
where $\eta$ is a gain, $c_{max}$ caps the cut at a fraction of the base release, and $\epsilon_t$ is multiplicative supply noise. Because the release depends on the teammates' conversion rates through $H$ and on the team stockpile through the cut, observing the supplier reveals exactly the quantities required by Equations~\ref{eqE1} and~\ref{eqE2}, while individual teammate rates remain hidden.

\end{appendices}

\newpage
\section{Extended Data}

\begin{figure}[h]
\centering
\includegraphics[width=0.95\textwidth]{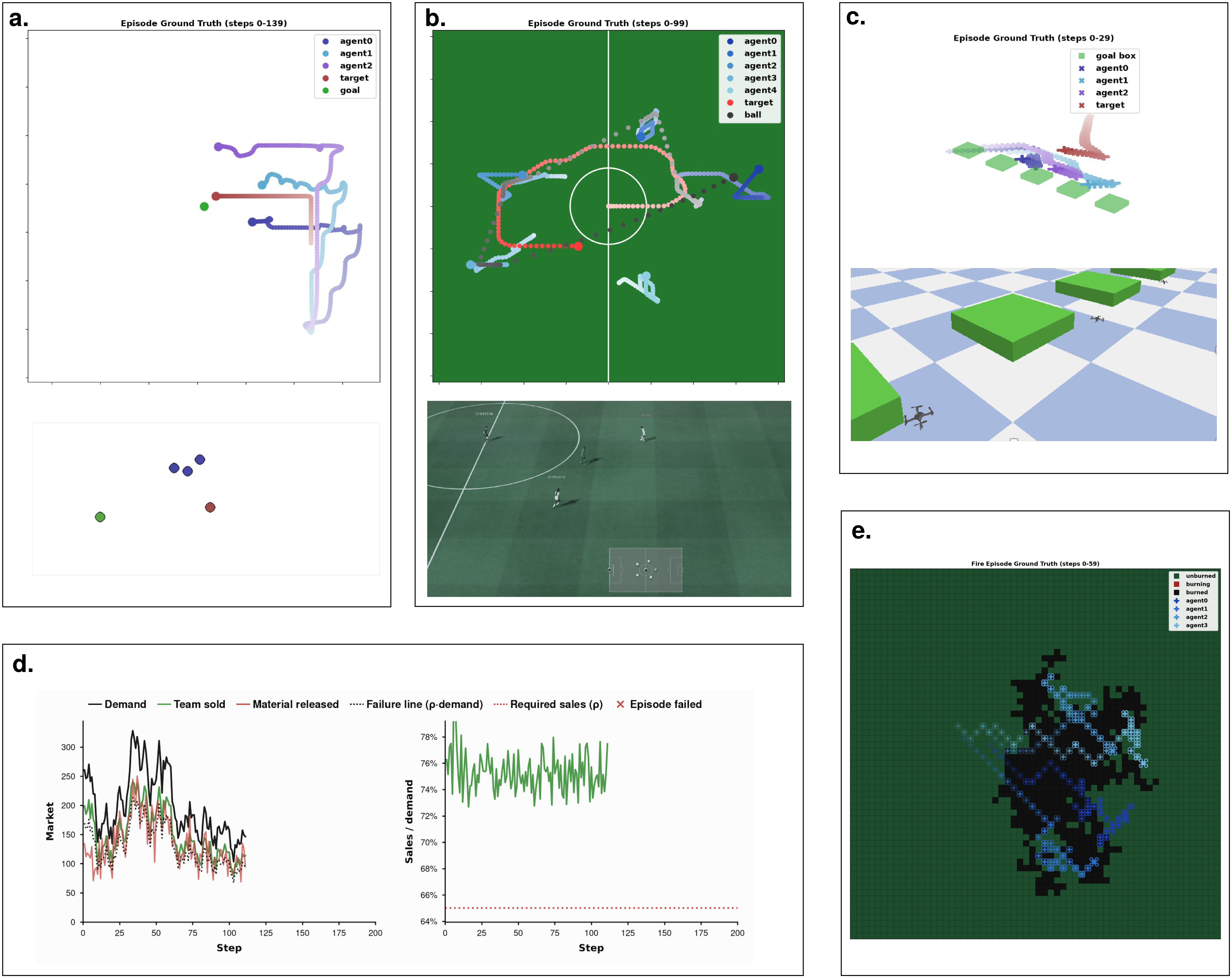}
\caption{An overlay image showing team behavior for each of the trained decentralized models over time. The figure includes: \textbf{a)} a successful predator prey episode (top left), and rendering of the MPE2 environment, \textbf{b)} an overlay plot of a Rondo Circle environment (top middle), and image of the rendered Google Research Football environment it was trained on, \textbf{c)} an overlay of part of a successful caravan defense episode (top right), and a rendering of the Pybullet-Drones environment trained on (bottom), \textbf{d)} an image from the market coordination environment (bottom left) showing both individual agents market production and the percentage of sales demand met, and \textbf{e)} an overlay of a successful fire control episode (bottom right).} \label{fig:environments}
\end{figure}

\bmhead{Data availability} The data generated and analyzed in this study is provided at: \url{https://github.com/cameronpm1/tells-results}, including instructions on how to regenerate rollouts for evaluation.

\bmhead{Code availability} The code for the environments, training and belief networks is available at: \url{https://github.com/cameronpm1/tells-train}

\bmhead{Acknowledgments} We would like to thank Dr. Erik Blasch at the Air Force Office of Scientific Research.

\bmhead{Funding}
This work was supported by the Air Force Office of Scientific Research.

\bmhead{Author contributions} C.M. and G.F. conceived the study. C.M. developed the allocentric belief framework, the training pipeline and the canonical correlation analysis. A.M. and H.G. contributed to the design and analysis of the predator-prey environment and drone caravan environment. A.K. assisted with building the market coordination environment. C.M. built all other environments including egocentric baselines, ran the experiments, and performed analysis on the data collected. C.M. wrote the manuscript with input from all authors. G.F. supervised the project and acquired funding. All authors reviewed and approved the final manuscript.

\bmhead{Competing interests} The authors declare no competing interests.

\bmhead{Additional information} Supplementary Information is available for this paper. Correspondence and requests for materials should be addressed to Cameron Mehlman.

\bibliography{sn-bibliography}

\end{document}